\documentclass[letterpaper,english,amscd,amsmath,amssymb,verbatim,floatfix,nofootinbib,eqsecnum]{article}
\usepackage[T1]{fontenc}
\usepackage[latin9]{inputenc}
\usepackage{babel}

\usepackage{graphicx}
\usepackage{amssymb}
\usepackage{esint}
\usepackage[unicode=true, pdfusetitle,
 bookmarks=true,bookmarksnumbered=false,bookmarksopen=false,
 breaklinks=false,pdfborder={0 0 1},backref=false,colorlinks=false]
 {hyperref}



\begin{document}

\title{Stochastic Vector Quantisers%
\thanks{This paper was submitted to a Special Issue of IEEE Trans. Information
Theory on Information-Theoretic Imaging on 30 July 1999. It was not
accepted for publication, but it underpins several subsequently published
papers.%
}}

\author{S P Luttrell}

\maketitle
\noindent \textbf{Abstract:} In this paper a stochastic generalisation
of the standard Linde-Buzo-Gray (LBG) approach to vector quantiser
(VQ) design is presented, in which the encoder is implemented as the
sampling of a vector of code indices from a probability distribution
derived from the input vector, and the decoder is implemented as a
superposition of reconstruction vectors, and the stochastic VQ is
optimised using a minimum mean Euclidean reconstruction distortion
criterion, as in the LBG case. Numerical simulations are used to demonstrate
how this leads to self-organisation of the stochastic VQ, where different
stochastically sampled code indices become associated with different
input subspaces. This property may be used to automate the process
of splitting high-dimensional input vectors into low-dimensional blocks
before encoding them.

\section{Introduction}

In vector quantisation a code book is used to encode each input vector
as a corresponding code index, which is then decoded (again, using
the codebook) to produce an approximate reconstruction of the original
input vector \cite{Gray1984,GershoGray1992}. The purpose of this
paper is to generalise the standard approach to vector quantiser (VQ)
design \cite{LindeBuzoGray1980}, so that each input vector is encoded
as a \textit{vector} of code indices that are stochastically sampled
from a probability distribution that depends on the input vector,
rather than as a \textit{single} code index that is the deterministic
outcome of finding which entry in a code book is closest to the input
vector. This will be called a stochastic VQ (SVQ), and it includes
the standard VQ as a special case. Note that this approach is different
from the various soft competition and stochastic relaxation schemes
that are used to train VQs (see e.g. \cite{YairZegerGersho1992}),
because here the probability distribution is an essential part of
the encoder, both during and after training.

One advantage of using the stochastic approach, which will be demonstrated
in this paper, is that it automates the process of splitting high-dimensional
input vectors into low-dimensional blocks before encoding them, because
minimising the mean Euclidean reconstruction error can encourage different
stochastically sampled code indices to become associated with different
input subspaces \cite{Luttrell1999a}. Another advantage is that it
is very easy to connect SVQs together, by using the vector of code
index probabilities computed by one SVQ as the input vector to another
SVQ \cite{Luttrell1999b}.

In section \ref{Sect:Theory} various pieces of previously published
theory are unified to give a coherent account of SVQs. In section
\ref{Sect:Simulations} the results of some new numerical simulations
are presented, which demonstrate how the code indices in a SVQ can
become associated in various ways with input subspaces.

\section{Theory}

\label{Sect:Theory}In this section various pieces of previously published
theory are unified to establish a coherent framework for modelling
SVQs.

In section \ref{Sect:FMC} the basic theory of folded Markov chains
(FMC) is given \cite{Luttrell1994}, and in section \ref{Sect:HighDimension}
it is extended to the case of high-dimensional input data \cite{Luttrell1997}.
In section \ref{Sect:AnalyticOptimum} some properties of the solutions
that emerge when the input vector lives on a 2-torus are summarised
\cite{Luttrell1999a}. Finally, in section \ref{Sect:ChainFMC} the
theory is further generalised to chains of linked FMCs \cite{Luttrell1999b}.

\subsection{Folded Markov Chains}

\label{Sect:FMC}The basic building block of the encoder/decoder model
used in this paper is the folded Markov chain (FMC) \cite{Luttrell1994}.
Thus an input vector $\mathbf{x}$ is encoded as a code index vector
$\mathbf{y}$, which is then subsequently decoded as a reconstruction
$\mathbf{x}^{\prime}$\ of the input vector. Both the encoding and
decoding operations are allowed to be probabilistic, in the sense
that $\mathbf{y}$ is a sample drawn from $\Pr\left(\mathbf{y}|\mathbf{x}\right)$,
and $\mathbf{x}^{\prime}$ is a sample drawn from $\Pr\left(\mathbf{x}^{\prime}|\mathbf{y}\right)$,
where $\Pr\left(\mathbf{y}|\mathbf{x}\right)$ and $\Pr\left(\mathbf{x}|\mathbf{y}\right)$
are Bayes' inverses of each other, as given by \begin{equation}
\Pr\left(\mathbf{x}|\mathbf{y}\right)=\frac{\Pr\left(\mathbf{y}|\mathbf{x}\right)\Pr\left(\mathbf{x}\right)}{\int d\mathbf{z}\Pr\left(\mathbf{y}|\mathbf{z}\right)\Pr\left(\mathbf{z}\right)}\label{Eq:Bayes}\end{equation}
and $\Pr\left(\mathbf{x}\right)$ is the prior probability from which
$\mathbf{x}$ was sampled.

\begin{figure}
\begin{centering}
\includegraphics[width=5cm]{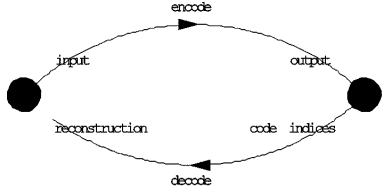}
\par\end{centering}

\caption{A folded Markov chain (FMC) in which an input vector $\mathbf{x}$
is encoded as a code index vector $\mathbf{y}$ that is drawn from
a conditional probability $\Pr\left(\mathbf{y}|\mathbf{x}\right)$,
which is then decoded as a reconstruction vector $\mathbf{x}^{\prime}$
drawn from the Bayes' inverse conditional probability $\Pr\left(\mathbf{x}^{\prime}|\mathbf{y}\right)$.}

\label{Fig:FMC}
\end{figure}

Because the chain of dependences in passing from $\mathbf{x}$ to
$\mathbf{y}$ and then to $\mathbf{x}^{\prime}$ is first order Markov
(i.e. it is described by the directed graph $\mathbf{x}\longrightarrow\mathbf{y}\longrightarrow\mathbf{x}^{\prime}$),
and because the two ends of this Markov chain (i.e. $\mathbf{x}$
and $\mathbf{x}^{\prime}$) live in the same vector space, it is called
a \textit{folded} Markov chain (FMC). The operations that occur in
an FMC are summarised in figure \ref{Fig:FMC}. %

In order to ensure that the FMC encodes the input vector optimally,
a measure of the reconstruction error must be minimised. There are
many possible ways to define this measure, but one that is consistent
with many previous results, and which also leads to many new results,
is the mean Euclidean reconstruction error measure $D$, which is
defined as \begin{equation}
D\equiv\int d\mathbf{x}\Pr\left(\mathbf{x}\right)\sum_{\mathbf{y}=1}^{\mathbf{M}}\Pr\left(\mathbf{y}|\mathbf{x}\right)\int d\mathbf{x}^{\prime}\Pr\left(\mathbf{x}^{\prime}|\mathbf{y}\right)\left\Vert \mathbf{x}-\mathbf{x}^{\prime}\right\Vert ^{2}\label{Eq:ObjectiveFMC}\end{equation}
where $\Pr\left(\mathbf{x}\right)\Pr\left(\mathbf{y}|\mathbf{x}\right)\Pr\left(\mathbf{x}^{\prime}|\mathbf{y}\right)$
is the joint probability that the FMC has state $\left(\mathbf{x},\mathbf{y},\mathbf{x}^{\prime}\right)$,
$\left\Vert \mathbf{x}-\mathbf{x}^{\prime}\right\Vert ^{2}$ is the
Euclidean reconstruction error, and $\int d\mathbf{x}\sum_{\mathbf{y}=1}^{\mathbf{M}}\int d\mathbf{x}^{\prime}\left(\cdots\right)$
sums over all possible states of the FMC (weighted by the joint probability).
The code index vector $\mathbf{y}$ is assumed to lie on a rectangular
lattice of size $\mathbf{M}$.

The Bayes' inverse probability $\Pr\left(\mathbf{x}^{\prime}|\mathbf{y}\right)$
may be integrated out of this expression for $D$ to yield \begin{equation}
D=2\int d\mathbf{x}\Pr\left(\mathbf{x}\right)\sum_{\mathbf{y}=1}^{\mathbf{M}}\Pr\left(\mathbf{y}|\mathbf{x}\right)\left\Vert \mathbf{x}-\mathbf{x}^{\prime}\left(\mathbf{y}\right)\right\Vert ^{2}\label{Eq:ObjectiveFMC2}\end{equation}
where the reconstruction vector $\mathbf{x}^{\prime}\left(\mathbf{y}\right)$
is defined as $\mathbf{x}^{\prime}\left(\mathbf{y}\right)\equiv\int d\mathbf{x}\Pr\left(\mathbf{x}|\mathbf{y}\right)\,\mathbf{x}$.
Because of the quadratic form of the objective function, it turns
out that $\mathbf{x}^{\prime}\left(\mathbf{y}\right)$ may be treated
as a free parameter whose optimum value (i.e. the solution of $\frac{\partial D}{\partial\mathbf{x}^{\prime}\left(\mathbf{y}\right)}=0$)
is $\int d\mathbf{x}\Pr\left(\mathbf{x}|\mathbf{y}\right)\,\mathbf{x}$,
as required.

If $D$ is now minimised with respect to the probabilistic encoder
$\Pr\left(\mathbf{y}|\mathbf{x}\right)$ and the reconstruction vector
$\mathbf{x}^{\prime}\left(\mathbf{y}\right)$, then the optimum has
the form \begin{eqnarray}
\Pr\left(\mathbf{y}|\mathbf{x}\right) & = & \delta_{\mathbf{y},\mathbf{y}\left(\mathbf{x}\right)}\nonumber \\
\mathbf{y}\left(\mathbf{x}\right) & = & \begin{array}{c}
\arg\min\\
\mathbf{y}\end{array}\left\Vert \mathbf{x}-\mathbf{x}^{\prime}\left(\mathbf{y}\right)\right\Vert ^{2}\nonumber \\
\mathbf{x}^{\prime}\left(\mathbf{y}\right) & = & \int d\mathbf{x}\Pr\left(\mathbf{x}|\mathbf{y}\right)\,\mathbf{x}\label{Eq:OptimumFMC}\end{eqnarray}
where $\Pr\left(\mathbf{y}|\mathbf{x}\right)$ has reduced to a deterministic
encoder (as described by the Kronecker delta $\delta_{\mathbf{y},\mathbf{y}\left(\mathbf{x}\right)}$),
$\mathbf{y}\left(\mathbf{x}\right)$ is a nearest neighbour encoding
algorithm using code vectors $\mathbf{x}^{\prime}\left(\mathbf{y}\right)$
to partition the input space into code cells, and (in an optimised
configuration) the $\mathbf{x}^{\prime}\left(\mathbf{y}\right)$ are
the centroids of these code cells. This is equivalent to a standard
VQ \cite{LindeBuzoGray1980}.

An extension of the standard VQ to the case of where the code index
is transmitted along a noisy communication channel before the reconstruction
is attempted \cite{Kumazawa1984,Farvardin1990} was derived in \cite{Luttrell1994},
and was shown to lead to a good approximation to a topographic mapping
neural network \cite{Kohonen1984}.

\subsection{High Dimensional Input Spaces}

\label{Sect:HighDimension}A problem with the standard VQ is that
its code book grows exponentially in size as the dimensionality of
the input vector is increased, assuming that the contribution to the
reconstruction error from each input dimension is held constant. This
means that such VQs are useless for encoding extremely high dimensional
input vectors, such as images. The usual solution to this problem
is to partition the input space into a number of lower dimensional
subspaces (or blocks), and then to encode each of these subspaces
separately. However, this produces an undesirable side effect, where
the boundaries of the blocks are clearly visible in the reconstruction;
this is the origin of the blocky appearance of reconstructed images,
for instance.

There is also a much more fundamental objection to this type of partitioning,
because the choice of blocks should ideally be decided in such a way
that the correlations \textit{within} a block are much stronger than
the correlations \textit{between} blocks. In the case of image encoding,
this will usually be the case if the partitioning is that each block
consists of contiguous image pixels. However, more generally, there
is no guarantee that the input vector statistics will respect the
partitioning in this convenient way. Thus it will be necessary to
deduce the best choice of blocks from the training data.

In order to solve the problem of finding the best partitioning, consider
the following constraint on $\Pr\left(\mathbf{y}|\mathbf{x}\right)$
and $\mathbf{x}^{\prime}\left(\mathbf{y}\right)$ in the FMC objective
function in equation \ref{Eq:ObjectiveFMC2} \begin{eqnarray}
\mathbf{y} & = & \left(y_{1},y_{2},\cdots,y_{n}\right),1\leq y_{i}\leq M\nonumber \\
\Pr\left(\mathbf{y}|\mathbf{x}\right) & = & \Pr\left(y_{1}|\mathbf{x}\right)\Pr\left(y_{2}|\mathbf{x}\right)\cdots\Pr\left(y_{n}|\mathbf{x}\right)\nonumber \\
\mathbf{x}^{\prime}\left(\mathbf{y}\right) & = & \frac{1}{n}\sum_{i=1}^{n}\mathbf{x}^{\prime}\left(y_{i}\right)\label{Eq:Approximation}\end{eqnarray}
Thus the code index vector $\mathbf{y}$ is assumed to be $n$-dimensional,
each component $y_{i}$ (for $i=1,2,\cdots,n$ and $1\leq y_{i}\leq M$)
is an independent sample drawn from $\Pr(y|\mathbf{x})$, and the
reconstruction vector $\mathbf{x}^{\prime}\left(\mathbf{y}\right)$
(vector argument)\ is assumed to be a superposition of $n$ contributions
$\mathbf{x}^{\prime}\left(y_{i}\right)$ (scalar argument) for $i=1,2,\cdots,n$.
As $D$ is minimised, this constraint allows partitioned solutions
to emerge by a process of self-organisation.

\begin{figure}
\begin{centering}
\includegraphics[width=5cm]{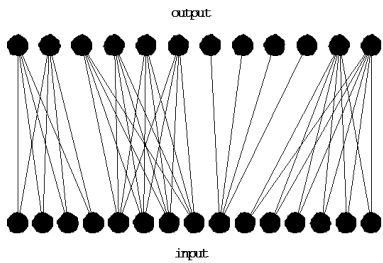}
\par\end{centering}

\caption{A typical solution in which the components of the input vector $\mathbf{x}$
and the range of values of the output code index $y$ are both partitioned
into blocks. These input and output blocks are connected together
as illustrated. More generally, the blocks may overlap each other.}

\label{Fig:Partitioning}
\end{figure}

For instance, solutions can have the structure illustrated in figure
\ref{Fig:Partitioning}. %
This type of structure is summarised as follows \begin{eqnarray}
\mathbf{x} & = & \left(\mathbf{x}_{1},\mathbf{x}_{2},\cdots,\mathbf{x}_{N}\right)\nonumber \\
\Pr\left(\mathbf{y}|\mathbf{x}\right) & = & \Pr\left(y_{1}|\mathbf{x}_{p\left(y_{1}\right)}\right)\Pr\left(y_{2}|\mathbf{x}_{p\left(y_{2}\right)}\right)\,\cdots\,\Pr\left(y_{n}|\mathbf{x}_{p\left(y_{n}\right)}\right)\label{Eq:ApproximationPartitioned}\end{eqnarray}
In this type of solution the input vector $\mathbf{x}$ is partitioned
as $\left(\mathbf{x}_{1},\mathbf{x}_{2},\cdots,\mathbf{x}_{N}\right)$,
the probability $\Pr\left(y|\mathbf{x}\right)$ reduces to $\Pr\left(y|\mathbf{x}_{p\left(y\right)}\right)$
which depends only on $\mathbf{x}_{p\left(y\right)}$, where the function
$p\left(y\right)$ computes the index of the block which code index
$y$ inhabits$.$ There is not an exact correspondence between this
type of partitioning and that used in the standard approach to encoding
image blocks, because here the $n$ code indices are spread at random
over the $N$ image blocks, which does not guarantee that every block
is encoded. Although, for a given $N$, if $n$ is chosen to be suffiently
large, then there is a virtual certainty that every block is encoded.
This is the price that has to be paid when it is assumed that the
code indices are drawn independently.

The constraints in equation \ref{Eq:Approximation} prevent the full
space of possible values of $\Pr\left(\mathbf{y}|\mathbf{x}\right)$
or $\mathbf{x}^{\prime}\left(\mathbf{y}\right)$ from being explored
as $D$ is minimised, so they lead to an \textit{upper bound} $D_{1}+D_{2}$
on the FMC objective function $D$ (i.e. $D\leq D_{1}+D_{2}$), which
may be derived as \cite{Luttrell1997}\begin{eqnarray}
D_{1} & \equiv & \frac{2}{n}\int d\mathbf{x}\Pr\left(\mathbf{x}\right)\sum_{y=1}^{M}\Pr\left(y|\mathbf{x}\right)\left\Vert \mathbf{x}-\mathbf{x}^{\prime}\left(y\right)\right\Vert ^{2}\nonumber \\
D_{2} & \equiv & \frac{2\left(n-1\right)}{n}\int d\mathbf{x}\Pr\left(\mathbf{x}\right)\left\Vert \mathbf{x}-\sum_{y=1}^{M}\Pr\left(y|\mathbf{x}\right)\mathbf{x}^{\prime}\left(y\right)\right\Vert ^{2}\label{Eq:D1D2}\end{eqnarray}
Note that $M$ and $n$ are model order parameters, whose values need
to be chosen appropriately for each encoder optimisation problem.

For $n=1$ only the $D_{1}$ term contributes, and it is equivalent
to the FMC objective function $D$ in equation \ref{Eq:ObjectiveFMC2}
with the vector code index $\mathbf{y}$ replaced by a scalar code
index $y$, so its minimisation leads to a standard vector quantiser
as in equation \ref{Eq:OptimumFMC}, in which each input vector is
approximated by a \textit{single} reconstruction vector $\mathbf{x}^{\prime}\left(y\right)$.

When $n$ becomes large enough that $D_{2}$ dominates over $D_{1}$,
the optimisation problem reduces to minimisation of the mean Euclidean
reconstruction error (approximately). This encourages the approximation
$\mathbf{x}\approx\sum_{y=1}^{M}\Pr\left(y|\mathbf{x}\right)\mathbf{x}^{\prime}\left(y\right)$
to hold, in which the input vector $\mathbf{x}$ is approximated as
a weighted (using weights $\Pr\left(y|\mathbf{x}\right)$) sum of
\textit{many} reconstruction vectors $\mathbf{x}^{\prime}\left(y\right)$.
In numerical simulations this has invariably led to solutions which
are a type of principal components analysis (PCA) of the input vectors,
where the expansion coefficients $\Pr\left(y|\mathbf{x}\right)$ are
constrained to be non-negative and sum to unity. Also, the approximation
$\mathbf{x}\approx\sum_{y=1}^{M}\Pr\left(y|\mathbf{x}\right)\mathbf{x}^{\prime}\left(y\right)$
is very good for solutions in which $\Pr\left(y|\mathbf{x}\right)$
depends on the whole of $\mathbf{x}$, rather than merely on a subspace
of $\mathbf{x}$, so this does not lead to a partitioned solution.

For intermediate values of $n$, where both $D_{1}$ and $D_{2}$
are comparable in size, partitioned solutions can emerge. However,
in this intermediate region the properties of the optimum solution
depend critically on the interplay between the statistical properties
of the training data and the model order parameters $M$ and $n$.
To illustrate how the choice of $M$ and $n$ affects the solution,
the case of input vectors that live on a 2-torus is summarised in
section \ref{Sect:AnalyticOptimum}.

When the full FMC objective function in equation \ref{Eq:ObjectiveFMC2}
is optimised it leads to the standard (deterministic) VQ in equation
\ref{Eq:OptimumFMC}. However, it turns out that the constrained FMC
objective function $D_{1}+D_{2}$ in equation \ref{Eq:D1D2} does
not allow a deterministic VQ to emerge (except in the case $n=1$),
because a more accurate reconstruction can be obtained by allowing
more than one code index to be sampled for each input vector. Because
of this behaviour, in which the encoder is stochastic both during
and after training, this type of constrained FMC will be called a
SVQ.

\subsection{Example: 2-Torus Case}

\label{Sect:AnalyticOptimum}A scene is defined as a number of objects
at specified positions and orientations, so is it specified by a low-dimensional
vector of scene coordinates, which are effectively the intrinsic coordinates
of a low-dimensional manifold. An image of that scene is an embedding
of the low-dimensional manifold in the high-dimensional space of image
pixel values. Because the image pixel values are non-linearly related
to the vector of scene coordinates, this embedding operation distorts
the manifold so that it becomes curved. The problem of finding the
optimal way to encode images may thus be viewed as the problem of
finding the optimal way to encode curved manifolds, where the instrinsic
dimensionality of the manifold is the same as the dimensionality of
the vector of scene coordinates.

The simplest curved manifold is the circle, and the next most simple
curved manifold is the 2-torus (which has 2 intrinsic circular coordinates).
By making extensive use of the fact that the optimal form of $\Pr\left(y|\mathbf{x}\right)$
must be a piecewise linear function of the input vector $\mathbf{x}$
\cite{Luttrell1999b}, the optimal encoders for these manifolds have
been derived analytically \cite{Luttrell1999a}. The toroidal case
is very interesting because it demonstrates the transition between
the unpartitioned and partitioned optimum solutions as $n$ is increased.
A 2-torus is a realistic model of the manifold generated by the linear
superposition of 2 sine waves, having fixed amplitudes and wavenumbers.
The phases of the 2 sine waves are then the 2 intrinsic circular coordinates
of this manifold, and if these phases are both uniformly distributed
on the circle, then $\Pr\left(y|\mathbf{x}\right)$ defines a constant
probability density on the 2-torus.

\begin{figure}
\begin{centering}
\includegraphics[width=5cm]{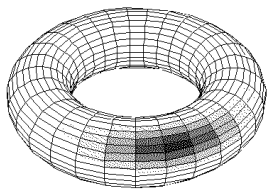}
\par\end{centering}

\caption{A typical probability $\Pr\left(y|\mathbf{x}\right)$ (only one $y$
value is illustrated)\ for encoding a 2-torus using a small value
of $n$. This defines a smoothly tapered localised region on the 2-torus.
The toroidal mesh serves only to help visualise the 2-torus.}

\label{Fig:TorusJoint}
\end{figure}

A typical $\Pr\left(y|\mathbf{x}\right)$ for small $n$ is illustrated
in figure \ref{Fig:TorusJoint}. %
 Only in the case where $n=1$ would $\Pr\left(y|\mathbf{x}\right)$
correspond to a sharply defined code cell; for $n>1$ the edges of
the code cells are tapered so that they overlap with one other. The
2-torus is covered with a large number of these overlapping code cells,
and when a code index $y$ is sampled from $\Pr\left(y|\mathbf{x}\right)$,
it allows a reconstruction of the input to be made to within an uncertainty
area commensurate with the size of a code cell. This type of encoding
is called \textit{joint} encoding, because the 2 intrinsic dimensions
of the 2-torus are \textit{simultaneously} encoded by $y$.

\begin{figure}
\begin{centering}
\includegraphics[width=5cm]{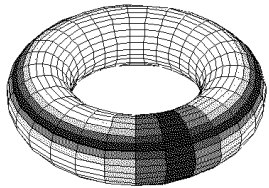}
\par\end{centering}

\caption{Two typical probabilities $\Pr\left(y_{1}|\mathbf{x}\right)$ and
$\Pr\left(y_{2}|\mathbf{x}\right)$ for encoding a 2-torus using a
large value of $n$. Each separately defines a smoothly tapered collar-shaped
region on the 2-torus. However, taken together, their region of intersection
defines a smoothly tapered localised region on the 2-torus. The toroidal
mesh serves only to help visualise the 2-torus.}

\label{Fig:TorusFactorial}
\end{figure}

A typical pair of $\Pr\left(y|\mathbf{x}\right)$ (i.e. $\Pr\left(y_{1}|\mathbf{x}\right)$
and $\Pr\left(y_{2}|\mathbf{x}\right)$)\ for large $n$\ is illustrated
in figure \ref{Fig:TorusFactorial}. %
The partitioning splits the code indices into two different types:
one type encodes one of the intrinsic dimensions of the 2-torus, and
the other type encodes the other intrinsic dimension of the 2-torus,
so each code cell is a tapered collar-shaped region. When the code
indices\ $\left(y_{1},y_{2},\cdots,y_{n}\right)$ are sampled from
$\Pr\left(y|\mathbf{x}\right)$, they allow a reconstruction of the
input to be made to within an uncertainty area commensurate with the
size of the region of intersection of a pair of orthogonal code cells,
as illustrated in figure \ref{Fig:TorusFactorial}. This type of encoding
is called \textit{factorial} encoding, because the 2 intrinsic dimensions
of the 2-torus are \textit{separately} encoded in $\left(y_{1},y_{2},\cdots,y_{n}\right)$.

For a 2-torus there is an upper limit $M\approx12$ beyond which the
optimum solution is always a joint encoder (as shown in figure \ref{Fig:TorusJoint}).
This limit arises because when $M\gtrsim12$ the code book is sufficiently
large that the joint encoder gives the best reconstruction for all
values of $n$. This result critically depends on the fact that as
$n$ is increased the code cells overlap progressively more and more,
so the accuracy of the reconstruction progressively increases. For
$M\gtrsim12$ the rate at which the accuracy of the joint encoder
improves (as $n$ increases) is sufficiently great that it is always
better than that of the factorial encoder (which also improves as
$n$ increases).

\subsection{Chains of Linked FMCs}

\label{Sect:ChainFMC}Thus far it has been shown that the FMC objective
function in equation \ref{Eq:ObjectiveFMC2}, with the constraints
imposed in equation \ref{Eq:Approximation}, leads to useful properties,
such as the automatic partitioning of the code book to yield the factorial
encoder, such as that illustrated in figure \ref{Fig:TorusFactorial}
(and more generally, as illustrated in figure \ref{Fig:Partitioning}).
The free parameters $\left(M,n\right)$ (i.e. the size of the code
book, and the number of code indices sampled)\ can be adjusted to
obtain an optimal solution that has the desired properties (e.g. a
joint or a factorial encoder, as in figures \ref{Fig:TorusJoint}
and \ref{Fig:TorusFactorial}, respectively). However, since there
are only 2 free parameters, there is a limit to the variety of types
of properties that the optimal solution can have. It would thus be
very useful to introduce more free parameters.

\begin{figure}
\begin{centering}
\includegraphics[width=5cm]{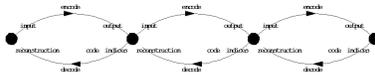}
\par\end{centering}

\caption{A chain of linked FMCs, in which the output from each stage is its
vector of posterior probabilities (for all values of the code index),
which is then used as the input to the next stage. Only 3 stages are
shown, but any number may be used. More generally, any acyclically
linked network of FMCs may be used.}

\label{Fig:FMCchain}
\end{figure}

The FMC\ illustrated in figure \ref{Fig:FMC} may be generalised
to a chain of linked FMCs as shown in figure \ref{Fig:FMCchain}.
Each stage in this chain is an FMC of the type shown in figure \ref{Fig:FMC},
and the vector of probabilities (for all values of the code index)
computed by each stage is used as the input vector to the next stage;
there are other ways of linking the stages together, but this is the
simplest possibility. The overall objective function is a weighted
sum of the FMC objective functions derived from each stage. The total
number of free parameters in an $L$ stage chain is $3L-1$, which
is the sum of 2 free parameters for each of the $L$ stages, plus
$L-1$ weighting coefficients; there are $L-1$ rather than $L$ weighting
coefficients because the overall normalisation of the objective function
does not affect the optimum solution.

The chain of linked FMCs may be expressed mathematically by first
of all introducing an index $l$ to allow different stages of the
chain to be distinguished thus \begin{eqnarray}
M & \longrightarrow & M^{\left(l\right)},y\longrightarrow y^{\left(l\right)}\nonumber \\
\mathbf{x} & \longrightarrow & \mathbf{x}^{\left(l\right)},\mathbf{x}^{\prime}\longrightarrow\mathbf{x}^{\left(l\right)\prime}\nonumber \\
n & \longrightarrow & n^{\left(l\right)},D\longrightarrow D^{\left(l\right)}\nonumber \\
D_{1} & \longrightarrow & D_{1}^{\left(l\right)},D_{2}\longrightarrow D_{2}^{\left(l\right)}\label{Eq:ChainNotation}\end{eqnarray}
The stages are then defined and linked together thus (the detailed
are given only as far as the input to the third stage) \begin{eqnarray}
\mathbf{x}^{\left(1\right)} & \longrightarrow & y^{\left(1\right)}\longrightarrow\mathbf{x}^{\left(1\right)\prime}\nonumber \\
\mathbf{x}^{\left(2\right)} & = & \left(x_{1}^{\left(2\right)},x_{2}^{\left(2\right)},\cdots,x_{M^{\left(1\right)}}^{\left(2\right)}\right)\nonumber \\
x_{i}^{\left(2\right)} & = & \Pr\left(y^{\left(1\right)}=i|\mathbf{x}^{\left(1\right)}\right),1\leq i\leq M^{\left(1\right)}\nonumber \\
\mathbf{x}^{\left(2\right)} & \longrightarrow & y^{\left(2\right)}\longrightarrow\mathbf{x}^{\left(2\right)\prime}\nonumber \\
\mathbf{x}^{\left(3\right)} & = & \left(x_{1}^{\left(3\right)},x_{2}^{\left(3\right)},\cdots,x_{M^{\left(2\right)}}^{\left(3\right)}\right)\nonumber \\
x_{i}^{\left(3\right)} & = & \Pr\left(y^{\left(2\right)}=i|\mathbf{x}^{\left(2\right)}\right),1\leq i\leq M^{\left(2\right)}\label{Eq:ChainLink}\end{eqnarray}
The objective function and its upper bound are then given by \begin{eqnarray}
D & = & \sum_{l=1}^{L}s^{\left(l\right)}D^{\left(l\right)}\nonumber \\
 & \leq & D_{1}+D_{2}\nonumber \\
 & = & \sum_{l=1}^{L}s^{\left(l\right)}\left(D_{1}^{\left(l\right)}+D_{2}^{\left(l\right)}\right)\label{Eq:ChainObjective}\end{eqnarray}
where $s^{\left(l\right)}\geq0$ is the weighting that is applied
to the contribution of stage $l$ of the chain to the overall objective
function.

The piecewise linearity property enjoyed by $\Pr\left(y|\mathbf{x}\right)$
in a single stage chain also holds for all of the $\Pr\left(y^{\left(l\right)}|\mathbf{x}^{\left(l\right)}\right)$
in a multi-stage chain, provided that the stages are linked together
as prescribed in equation \ref{Eq:ChainLink} \cite{Luttrell1999b}.
This will allow optimum analytic solutions to be derived by an extension
of the single stage methods used in \cite{Luttrell1999a}.

\section{Simulations}

\label{Sect:Simulations}In this section the results of various numerical
simulations are presented, which demonstrate some of the types of
behaviour exhibited by an encoder that consists of a chain of linked
FMCs. Synthetic, rather than real, training data are used in all of
the simulations, because this allows the basic types of behaviour
to be cleanly demonstrated.

In section \ref{Sect:TrainingAlgorithm} the training algorithm is
presented. In section \ref{Sect:TrainingData} the training data is
described. In section \ref{Sect:IndependentTargets} a single stage
encoder is trained on data that is a superposition of two randomly
positioned objects. In section \ref{Sect:CorrelatedTargets} this
is generalised to objects with correlated positions, and three different
types of behaviour are demonstrated: factorial encoding using both
a 1-stage and a 2-stage encoders (section \ref{Sect:FactorialEncoding}),
joint encoding using a 1-stage encoder (section \ref{Sect:JointEncoding}),
and invariant encoding (i.e. ignoring a subspace of the input space
altogether) using a 2-stage encoder (section \ref{Sect:InvariantEncoding}).

\subsection{Training Algorithm}

\label{Sect:TrainingAlgorithm}Assuming that $\Pr\left(y|\mathbf{x}\right)$
is modelled as in appendix \ref{Sect:DerivativeObjectiveFunction}
(i.e. $\Pr\left(y|\mathbf{x}\right)=\frac{Q\left(y|\mathbf{x}\right)}{\sum_{y^{\prime}=1}^{M}Q\left(y^{\prime}|\mathbf{x}\right)}$
and $Q\left(y|\mathbf{x}\right)=\frac{1}{1+\exp\left(-\mathbf{w}\left(y\right)\cdot\mathbf{x}-b\left(y\right)\right)}$),
then the partial derivatives of $D_{1}+D_{2}$ with respect to the
3 types of parameters in a single stage of the encoder may be denoted
as \begin{eqnarray}
\mathbf{g}_{w}\left(y\right) & \equiv & \frac{\partial\left(D_{1}+D_{2}\right)}{\partial\mathbf{w}\left(y\right)}\nonumber \\
g_{b}\left(y\right) & \equiv & \frac{\partial\left(D_{1}+D_{2}\right)}{\partial b\left(y\right)}\nonumber \\
\mathbf{g}_{x}\left(y\right) & \equiv & \frac{\partial\left(D_{1}+D_{2}\right)}{\partial\mathbf{x}^{\prime}\left(y\right)}\end{eqnarray}
This may be generalised to each stage of a multi-stage encoder by
including an $\left(l\right)$ superscript, and ensuring that for
each stage the partial derivatives include the additional contributions
that arise from forward propagation through later stages; this is
essentially an application of the chain rule of differentiation, using
the derivatives $\frac{\partial\mathbf{x}^{\left(l+1\right)}}{\partial\mathbf{w}^{\left(l\right)}\left(y^{\left(l\right)}\right)}$
and $\frac{\partial\mathbf{x}^{\left(l+1\right)}}{\partial b^{\left(l\right)}\left(y^{\left(l\right)}\right)}$
to link the stages together (see appendix \ref{Sect:DerivativeObjectiveFunction}).

A\ simple algorithm for updating these parameters is (omitting the
$\left(l\right)$ superscript, for clarity) \begin{eqnarray}
\mathbf{w}\left(y\right) & \longrightarrow & \mathbf{w}\left(y\right)-\varepsilon\,\frac{\mathbf{g}_{w}\left(y\right)}{g_{w,0}}\nonumber \\
b\left(y\right) & \longrightarrow & b\left(y\right)-\varepsilon\,\frac{g_{b}\left(y\right)}{g_{b,0}}\nonumber \\
\mathbf{x}^{\prime}\left(y\right) & \longrightarrow & \mathbf{x}^{\prime}\left(y\right)-\varepsilon\,\frac{\mathbf{g}_{x}\left(y\right)}{g_{x,0}}\end{eqnarray}
where $\varepsilon$ is a small update step size parameter, and the
three normalisation factors are defined as \begin{eqnarray}
g_{w,0} & \equiv & \begin{array}{c}
\max\\
y\end{array}\sqrt{\frac{\left\Vert \mathbf{g}_{w}\left(y\right)\right\Vert ^{2}}{\dim\mathbf{x}}}\nonumber \\
g_{b,0} & \equiv & \begin{array}{c}
\max\\
y\end{array}\left|b\left(y\right)\right|\nonumber \\
g_{x,0} & \equiv & \begin{array}{c}
\max\\
y\end{array}\sqrt{\frac{\left\Vert \mathbf{g}_{x}\left(y\right)\right\Vert ^{2}}{\dim\mathbf{x}}}\end{eqnarray}
The $\frac{\mathbf{g}_{w}\left(y\right)}{g_{w,0}}$ and $\frac{\mathbf{g}_{x}\left(y\right)}{g_{x,0}}$
factors ensure that the maximum update step size for $\mathbf{w}\left(y\right)$
and $\mathbf{x}^{\prime}\left(y\right)$ is $\varepsilon\dim\mathbf{x}$
(i.e. $\varepsilon$ per dimension), and the $\frac{g_{b}\left(y\right)}{g_{b,0}}$
factor ensures that the maximum update step size for $b\left(y\right)$
is $\varepsilon$. This update algorithm can be generalised to use
a different $\varepsilon$ for each stage of the encoder, and also
to allow a different $\varepsilon$ to be used for each of the 3 types
of parameter. Furthermore, the size of $\varepsilon$ can be varied
as training proceeds, usually starting with a large value, and then
gradually reducing its size as the solution converges. It is not possible
to give general rules for exactly how to do this, because training
conditions depend very much on the statistical properties of the training
set.

\subsection{Training Data}

\label{Sect:TrainingData}The key property that this type of self-organising
encoder exhibits is its ability to automatically split up high-dimensional
input spaces into lower-dimensional subspaces, each of which is separately
encoded. For instance, see section \ref{Sect:HighDimension} for a
summary of the analytically solved case of training data that lives
on a simple curved manifold (i.e. a 2-torus). This self-organisation
manifests itself in many different ways, depending on the interplay
between the statistical properties of the training data, and the $3$
free parameters (i.e. the code book size $M$, the number of code
indices sampled $n$, and the stage weighting $s$)\ per stage of
the encoder (see section \ref{Sect:ChainFMC}). However, it turns
out that the joint and factorial encoders (of the same general type
as those obtained in the case of a 2-torus) are also the optimum solutions
for more general curved manifolds.

\begin{figure}
\begin{centering}
\includegraphics[width=5cm]{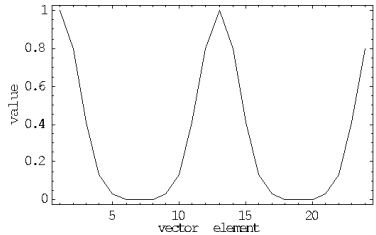}
\par\end{centering}

\caption{An example of a typical training vector for $M=24$. Each object is
a Gaussian hump with a half-width of 1.5 units, and peak amplitude
of 1. The overall input vector is formed as a linear superposition
of the 2 objects. Note that the input vector is wrapped around circularly
to remove minor edge effects that would otherwise arise.}

\label{Fig:ExampleData}
\end{figure}

In order to demonstrate the various different basic types of self-organisation
it is necessary to use synthetic training data with controlled properties.
All of the types of self-organisation that will be demonstrated in
this paper may be obtained by training a 1-stage or 2-stage encoder
on 24-dimensional data (i.e. $M=24$) that consists of a superposition
of a pair of identical objects (with circular wraparound to remove
edge effects), such as is shown in figure \ref{Fig:ExampleData}.

The training data is thus uniformly distributed on a manifold with
2 intrinsic circular coordinates, which is then embedded in a 24-dimensional
image space. The embedding is a curved manifold, but is \textit{not}
a 2-torus, and there are two reasons for this. Firstly, even though
the manifold has 2 intrinsic circular coordinates, the non-linear
embedding distorts these circles in the 24-dimensional embedding space
so that they are not planar (i.e. the profile of each object lives
in the \textit{full} 24-dimensional embedding space). Secondly, unlike
a 2-torus, each point on the manifold maps to itself under interchange
of the pair of circular coordinates, so the manifold is covered twice
by a 2-torus (i.e. the objects are identical, so it makes no difference
if they are swapped over). However, these differences do not destroy
the general character of the joint and factorial encoder solutions
that were obtained in section \ref{Sect:HighDimension}.

In the simulations presented below, two different methods of selecting
the object positions are used: either the positions are statistically
independent, or they are correlated. In the independent case, each
object position is a random integer in the interval $\left[1,24\right]$.
In the correlated case, the first object position is a random integer
in the interval $\left[1,24\right]$, and the second object position
is chosen \textit{relative to} the first one as an integer in the
range $\left[4,8\right]$, so that the mean object separation is $6$
units.

\subsection{Independent Objects}

\label{Sect:IndependentTargets}The simplest demonstration is to let
a single stage encoder discover the fact that the training data consists
of a superposition of a pair of objects, which is an example of independent
component analysis (ICA) or blind signal separation (BSS) \cite{Hyvarinen1999}.
This may readily be done by setting the parameters values as follows:
code book size $M=16$, number of code indices sampled $n=20$, $\varepsilon=0.2$
for $250$ training steps, $\varepsilon=0.1$ for a further $250$
training steps.

\begin{figure}
\begin{centering}
\includegraphics[width=5cm]{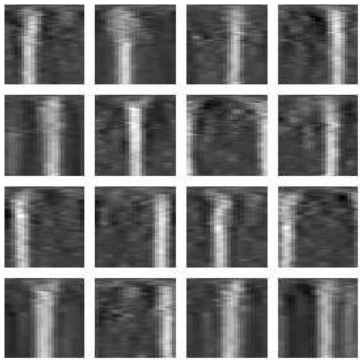}
\par\end{centering}

\caption{A factorial encoder emerges when a single stage encoder is trained
on data that is a superposition of 2 objects in independent locations.}

\label{Fig:IndependentTargets}
\end{figure}

The self-organisation of the $16$ reconstruction vectors as training
progresses (measured down the page) is shown in figure \ref{Fig:IndependentTargets}.

After some initial confusion, the reconstruction vectors self-organise
so that each code index corresponds to a \textit{single} object at
a well defined location. This behaviour is non-trivial, because each
training vector is a superposition of a \textit{pair} of objects at
independent locations, so typically more than one code index must
be sampled by the encoder, which is made possible by the relatively
large choice $n=20$. This result is a factorial encoder, because
the objects are encoded separately. This is a rudimentary example
of the type of solution that was illustrated in figure \ref{Fig:Partitioning},
although here the blocks overlap each other.

The case of a joint encoder requires a rather large code book when
the objects are independent. However, when correlations between the
objects are introduced then the code book can be reduced to a manageable
size, as will be demonstrated in the next section.

\subsection{Correlated Objects}

\label{Sect:CorrelatedTargets}
\begin{figure}
\begin{centering}
\includegraphics[width=5cm]{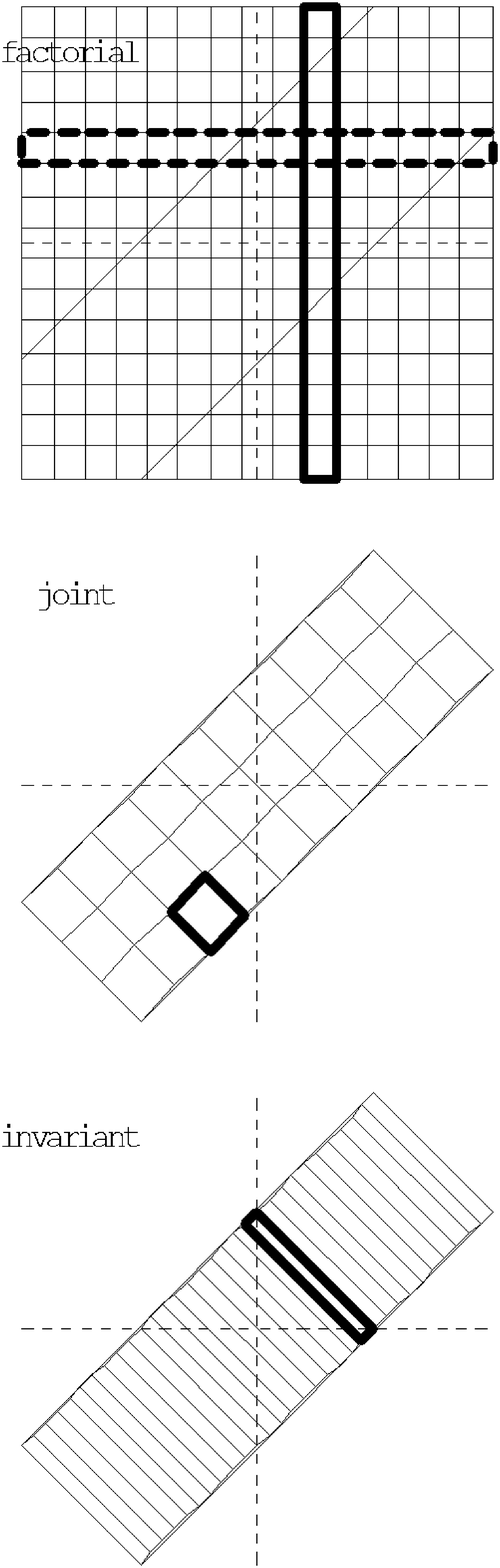}
\par\end{centering}

\caption{Three alternative ways of using $30$ code indices to encode a pair
of correlated variables. The\ typical code cells are shown in bold.}

\label{Fig:EncoderTypes}
\end{figure}

A more interesting situation arises
if the positions of the pair of objects are mutually correlated, so
that the training data is non-uniformly distributed on a manifold
with 2 intrinsic circular coordinates. The pair of objects can then
be encoded in 3 fundamentally different ways:
\begin{enumerate}
\item Factorial encoder. This encoder ignores the correlations between the
objects, and encodes them as if they were 2 independent objects. Each
code index would thus encode a single object position, so many code
indices must be sampled in order to virtually guarantee that both
object positions are encoded. This result is a type of independent
component analysis (ICA) \cite{Hyvarinen1999}.
\item Joint encoder. This encoder regards each possible joint placement
of the 2 objects as a distinct configuration. Each code index would
thus encode a pair of object positions, so only one code index needs
to be sampled in order to guarantee that both object positions are
encoded. This result is basically the same as what would be obtained
by using a standard VQ \cite{LindeBuzoGray1980}.
\item Invariant encoder. This encoder regards each possible placement of
the centroid of the 2 objects as a distinct configuration, but regards
all possible object separations (for a given centroid) as being equivalent.
Each code index would thus encode only the centroid of the pair of
objects. This type of encoder does not arise when the objects are
independent. This is similar to self-organising transformation invariant
detectors described in \cite{Webber1994}. 
\end{enumerate}
Each of these 3 possibilities is shown in figure \ref{Fig:EncoderTypes},
where the diagrams are meant only to be illustrative. The correlated
variables live in the large 2-dimensional rectangular region extending
from bottom-left to top-right of each diagram. For data of the type
shown in figure \ref{Fig:ExampleData}, the rectangular region is
in reality the curved manifold generated by varying the pair of object
coordinates, and the invariance of the data under interchange of the
pair of object coodinates means that the upper left and lower right
halves of each diagram cover the manifold twice. %

The factorial encoder has two orthogonal sets of long thin rectangular
code cells, and the diagram shows how a pair of such cells intersect
to define a small square code cell. The joint encoder behaves as a
standard vector quantiser, and is illustrated as having a set of square
code cells, although their shapes will not be as simple as this in
practice. The invariant encoder has a set of long thin rectangular
code cells that encode only the long diagonal dimension.

In all 3 cases there is overlap between code cells. In the case of
the factorial and joint encoders the overlap tends to be only between
nearby code cells, whereas in the case of an invariant encoder the
range of the overlap is usually much greater, as will be seen in the
numerical simulations below. In practice the optimum encoder may not
be a clean example of one of the types illustrated in figure \ref{Fig:EncoderTypes},
as will also be seen in the numerical simulations below.

\subsubsection{Factorial Encoding}

\label{Sect:FactorialEncoding}A factorial encoder may be trained
by setting the parameter values as follows: code book size $M=16$,
number of code indices sampled $n=20$, $\varepsilon=0.2$ for $500$
training steps, $\varepsilon=0.1$ for a further $500$ training steps.
This is the same as in the case of independent objects, except that
the number of training steps has been doubled.

\begin{figure}
\begin{centering}
\includegraphics[width=5cm]{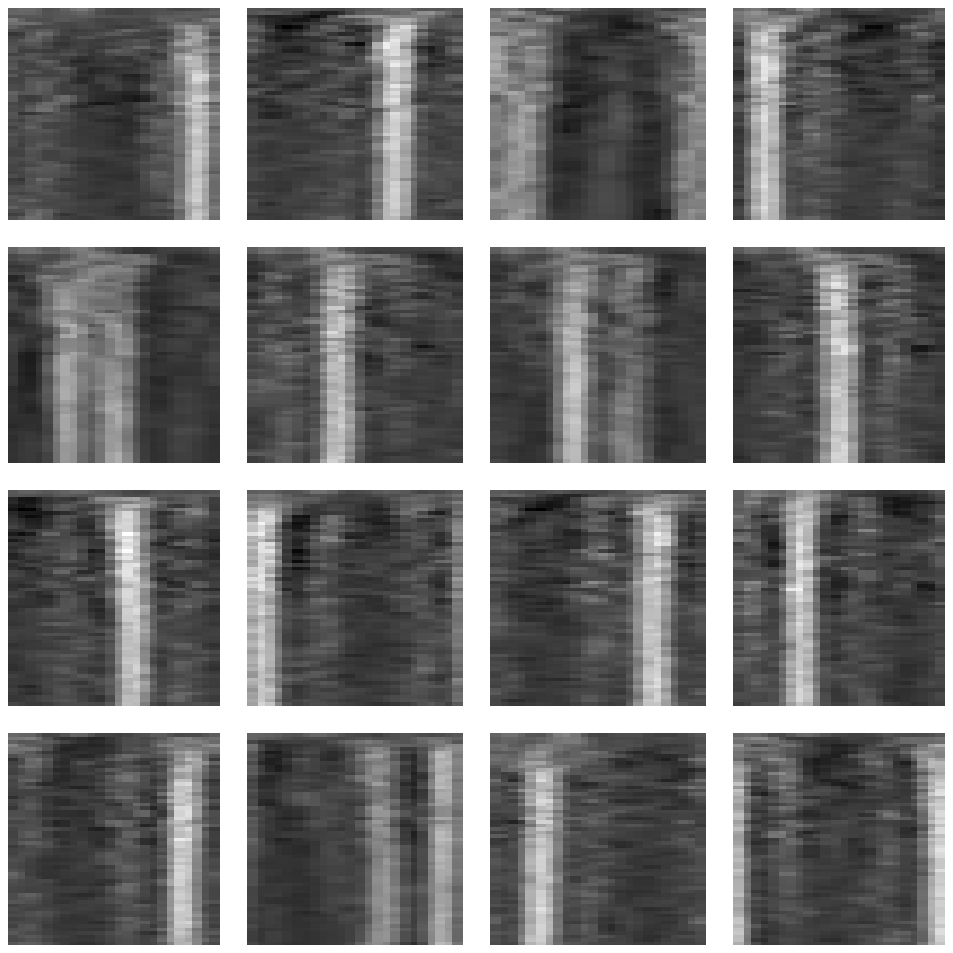}
\par\end{centering}

\caption{A factorial encoder emerges when a single stage encoder is trained
on data that is a superposition of 2 objects in correlated locations.}

\label{Fig:CorrelatedTargetsFactorial}
\end{figure}

The result is shown in figure \ref{Fig:CorrelatedTargetsFactorial}
which should be compared with the result for independent objects in
figure \ref{Fig:IndependentTargets}. The presence of correlations
degrades the quality of this factorial code relative to the case of
independent objects. The contamination of the factorial code takes
the form of a few code indices which respond jointly to the pair of
objects.

The joint coding contamination of the factorial code can be reduced
by using a 2-stage encoder, in which the second stage has the same
values of $M$ and $n$ as the first stage (although identical parameter
values are not necessary), and (in this case) both stages have the
same weighting in the objective function (see equation \ref{Eq:ChainObjective}).

\begin{figure}
\begin{centering}
\includegraphics[width=5cm]{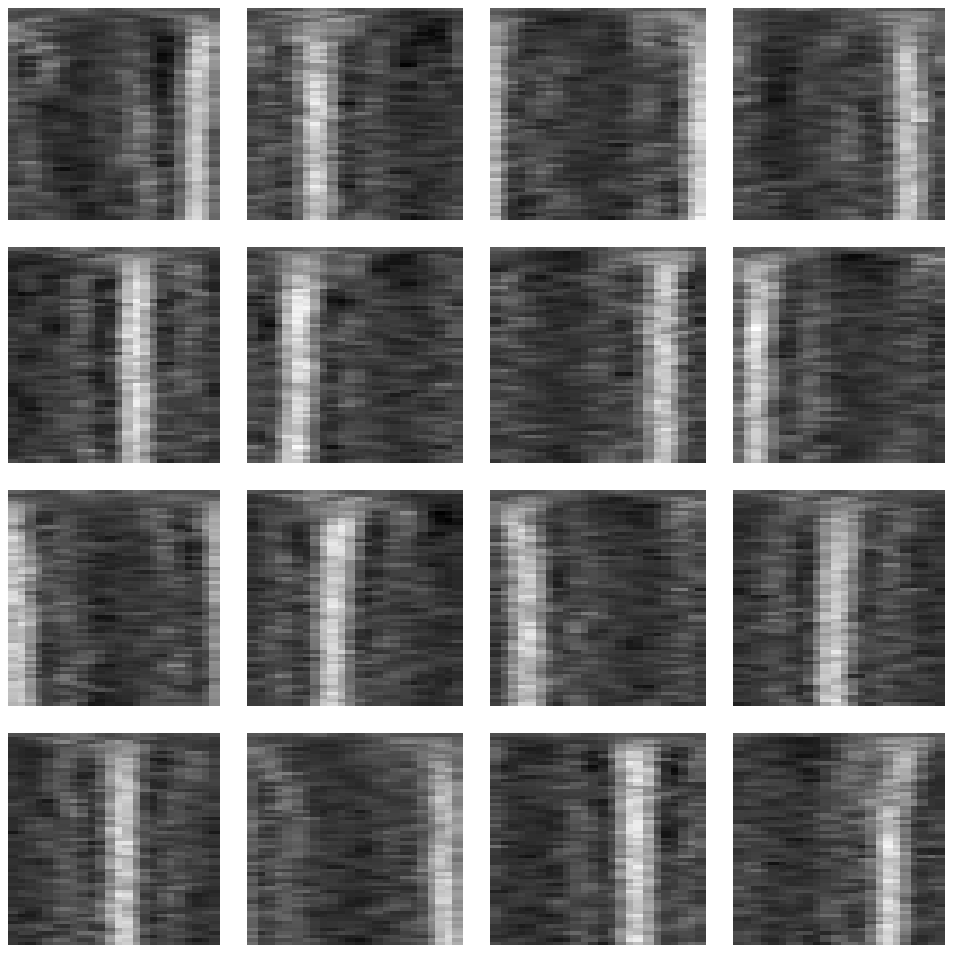}
\par\end{centering}

\caption{The factorial encoder is improved, by the removal of the joint encoding
contamination, when a 2-stage encoder is used.}

\label{Fig:CorrelatedTargetsFactorial2}
\end{figure}

The results are shown in figure \ref{Fig:CorrelatedTargetsFactorial2}.
The reason that the second stage encourages the first to adopt a pure
factorial code is quite subtle. The result shown in figure \ref{Fig:CorrelatedTargetsFactorial2}
will lead to the first stage producing an output in which 2 code indices
(one for each object) typically have probability $\frac{1}{2}$ of
being sampled, and all of the remaining code indices have a very small
probability (this is an approximation which ignores the fact that
the code cells overlap). On the other hand, figure \ref{Fig:CorrelatedTargetsFactorial}
will lead to an output in which the probability is sometimes concentrated
on a single code index. However, the contribution of the second stage
to the overall objective function encourages it to encode the vector
of probabilities output by the first stage with minimum Euclidean
reconstruction error, which is easier to do if the situation is as
in figure \ref{Fig:CorrelatedTargetsFactorial2} rather than as in
figure \ref{Fig:CorrelatedTargetsFactorial}. In effect, the second
stage likes to see an output from the first stage in which a large
number of code indices are each sampled with a low probability, which
favours factorial coding over joint encoding.

\subsubsection{Joint Encoding}

\label{Sect:JointEncoding}A joint encoder may be trained by setting
the parameter values as follows: code book size $M=16$, number of
code indices sampled $n=3$, $\varepsilon=0.2$ for $500$ training
steps, $\varepsilon=0.1$ for a further $500$ training steps, $\varepsilon=0.05$
for a further $1000$ training steps. This is the same as the parameter
values for the factorial encoder above, except that $n$ has been
reduced to $n=3$, and the training schedule has been extended.

\begin{figure}
\begin{centering}
\includegraphics[width=5cm]{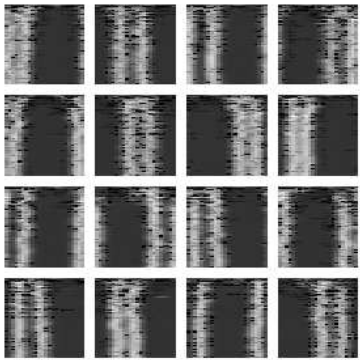}
\par\end{centering}

\caption{A joint encoder emerges when a single stage encoder is trained on
data that is a superposition of 2 objects in correlated locations.}

\label{Fig:CorrelatedTargetsJoint}
\end{figure}

The result is shown in figure \ref{Fig:CorrelatedTargetsJoint}. %
After some initial confusion, the reconstruction vectors self-organise
so that each code index corresponds to a \textit{pair} of objects
at well defined locations, so the code index jointly encodes the pair
of object positions; this is a joint encoder. The small value of $n$
prevents a factorial encoder from emerging.

\subsubsection{Invariant Encoding}

\label{Sect:InvariantEncoding}An invariant encoder may be trained
by using a 2-stage encoder, and setting the parameter values identically
in each stage as follows (where the weighting of the second stage
relative to the first is denoted as $s$): code book size $M=16$,
number of code indices sampled $n=3$, $\varepsilon=0.2$ and $s=5$
for $500$ training steps, $\varepsilon=0.1$ and $s=10$ for a further
$500$ training steps, $\varepsilon=0.05$ and $s=20$ for a further
$500$ training steps, $\varepsilon=0.05$ and $s=40$ for a further
$500$ training steps. This is basically the same as the parameter
values used for the joint encoder above, except that there are now
2 stages, and the weighting of the second stage is progressively increased
throughout the training schedule. Note that the large value that is
used for $s$ is offset to a certain extent by the fact that the ratio
of the normalisation of the inputs to the first and second stages
is very large; the anomalous normalisation of the input to the first
stage could be removed by insisting that the input to the first stage
is a vector of probabilities, but that is not done in these simulations.

\begin{figure}
\begin{centering}
\includegraphics[width=5cm]{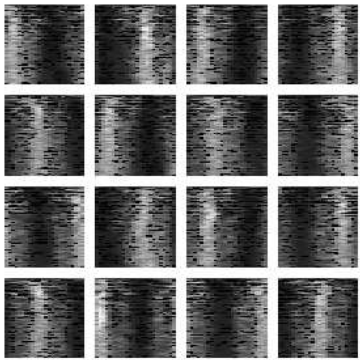}
\par\end{centering}

\caption{An invariant encoder emerges when 2-stage encoder is trained on data
that is a superposition of 2 objects in correlated locations.}

\label{Fig:CorrelatedTargetsInvariant}
\end{figure}

The result is shown in figure \ref{Fig:CorrelatedTargetsInvariant}.
During the early part of the training schedule the weighting of the
second stage is still relatively small, so it has the effect of turning
what would otherwise have been a joint encoder into a factorial encoder;
this is analogous to the effect observed when figure \ref{Fig:CorrelatedTargetsFactorial}
becomes figure \ref{Fig:CorrelatedTargetsFactorial2}. However, as
the training schedule progresses the weighting of the second stage
increases further, and the reconstruction vectors self-organise so
that each code index corresponds to a \textit{pair} of objects with
a well defined centroid but indeterminate separation. Thus each code
index encodes only the centroid of the pair of objects and ignores
their separation. This is a new type of encoder that arises when the
objects are correlated, and it will be called an \textit{invariant}
encoder, in recognition of the fact that its output is invariant with
respect to the separation of the objects.

Note that in these results there is a large amount of overlap between
the code cells, which should be taken into account when interpreting
the illustration in figure \ref{Fig:EncoderTypes}.

\section{Conclusions}

\label{Sect:Conclusions}The numerical results presented in this paper
show that a stochastic vector quantiser (VQ) can be trained to find
a variety of different types of way of encoding high-dimensional input
vectors. These input vectors are generated in two stages. Firstly,
a low-dimensional manifold is created whose intrinsic coordinates
are the positions of the objects in the scene; this corresponds to
generating the scene itself. Secondly, this manifold is non-linearly
embedded to create a curved manifold that lives in a high-dimensional
space of image pixel values; this corresponds to imaging the generated
scene.

Three fundamentally different types of encoder have been demonstrated,
which differ in the way that they build a reconstruction that approximates
the input vector:
\begin{enumerate}
\item A factorial encoder uses a reconstruction that is superposition of
a \textit{number} of vectors that each lives in a well defined input
subspace, which is useful for discovering constituent objects in the
input vector. This result is a type of independent component analysis
(ICA) \cite{Hyvarinen1999}.
\item A joint encoder uses a reconstruction that is a \textit{single} vector
that lives in the whole input space. This result is basically the
same as what would be obtained by using a standard VQ \cite{LindeBuzoGray1980}.
\item An invariant encoder uses a reconstruction that is a \textit{single}
vector that lives in a subspace of the whole input space, so it ignores
some dimensions of the input vector, which is therefore useful for
discovering correlated objects whilst rejecting uninteresting fluctuations
in their relative coordinates. This is similar to self-organising
transformation invariant detectors described in \cite{Webber1994}. 
\end{enumerate}
More generally, the encoder will be a hybrid of these basic types,
depending on the interplay between the statistical properties of the
input vector and the parameter settings of the SVQ.

\section{Acknowledgement}

I thank Chris Webber for many useful conversations that we had during
the course of this research.

\appendix

\section{Derivatives of the Objective Function}

\label{Sect:DerivativeObjectiveFunction}In order to minimise $D_{1}+D_{2}$
it is necessary to compute its derivatives. The derivatives were presented
in detail in \cite{Luttrell1997} for a single stage chain (i.e. a
single FMC). The purpose of this appendix is to extend this derivation
to a multi-stage chain of linked FMCs. In order to write the various
expressions compactly, infinitesimal variations will be used thoughout
this appendix, so that $\delta\left(uv\right)=\delta u\,\, v+u\,\delta v$
will be written rather than $\frac{\partial\left(uv\right)}{\partial\theta}=\frac{\partial u}{\partial\theta}\, v+u\,\frac{\partial v}{\partial\theta}$
(for some parameter $\theta$). The calculation will be done in a
top-down fashion, differentiating the objective function first, then
differentiating anything that the objective function depends on, and
so on following the dependencies down until only constants are left
(this is essentially the chain rule of differentiation).

The derivative of $D_{1}+D_{2}$ (defined in equation \ref{Eq:ChainObjective})
is given by

\begin{equation}
\delta\sum_{l=1}^{L}s^{\left(l\right)}\left(D_{1}^{\left(l\right)}+D_{2}^{\left(l\right)}\right)=\sum_{l=1}^{L}s^{\left(l\right)}\left(\delta D_{1}^{\left(l\right)}+\delta D_{2}^{\left(l\right)}\right)\label{Eq:DerivativeChainObjective}\end{equation}

The derivatives of the $D_{1}^{\left(l\right)}$ and $D_{2}^{\left(l\right)}$
parts (defined in equation \ref{Eq:D1D2}, with appropriate $\left(l\right)$
superscripts added) of the contribution of stage $l$ to $D_{1}+D_{2}$
are given by (dropping the $\left(l\right)$ superscripts again, for
clarity) \begin{eqnarray}
\delta D_{1} & = & \frac{2}{n}\int d\mathbf{x}\Pr\left(\mathbf{x}\right)\sum_{y=1}^{M}\left(\begin{array}{c}
\delta\Pr\left(y|\mathbf{x}\right)\left\Vert \mathbf{x}-\mathbf{x}^{\prime}\left(y\right)\right\Vert ^{2}\\
+2\Pr\left(y|\mathbf{x}\right)\left(\delta\mathbf{x-}\delta\mathbf{x}^{\prime}\left(y\right)\right)\cdot\left(\mathbf{x}-\mathbf{x}^{\prime}\left(y\right)\right)\end{array}\right)\nonumber \\
\delta D_{2} & = & \frac{4\left(n-1\right)}{n}\int d\mathbf{x}\Pr\left(\mathbf{x}\right)\sum_{y=1}^{M}\left[\begin{array}{c}
\left(\begin{array}{c}
\delta\mathbf{x}\\
-\sum_{y^{\prime}=1}^{M}\left(\begin{array}{c}
\delta\Pr\left(y^{\prime}|\mathbf{x}\right)\mathbf{x}^{\prime}\left(y^{\prime}\right)\\
+\Pr\left(y^{\prime}|\mathbf{x}\right)\delta\mathbf{x}^{\prime}\left(y^{\prime}\right)\end{array}\right)\end{array}\right)\\
\cdot\left(\mathbf{x}-\sum_{y^{\prime}=1}^{M}\Pr\left(y^{\prime}|\mathbf{x}\right)\mathbf{x}^{\prime}\left(y^{\prime}\right)\right)\end{array}\right]\label{Eq:DerivativeD1D2}\end{eqnarray}

In numerical simulations the exact piecewise linear solution for the
optimum $\Pr\left(y|\mathbf{x}\right)$ (see section \ref{Sect:AnalyticOptimum})\
will \textit{not} be sought, rather $\Pr\left(y|\mathbf{x}\right)$
will be modelled using a simple parametric form, and then the parameters
will be optimised. This model of $\Pr\left(y|\mathbf{x}\right)$ will
not in general include the ideal piecewise linear optimum solution,
so using it amounts to replacing $D_{1}+D_{2}$, which is an upper
bound on the objective function $D$ (see equation \ref{Eq:ChainObjective}),
by an even weaker upper bound on $D$. The justification for using
this approach rests on the quality of the results that are obtained
from the resulting numerical simulations (see section \ref{Sect:Simulations}).

The first step in modelling $\Pr\left(y|\mathbf{x}\right)$ is to
explicitly state the fact that it is a probability, which is a normalised
quantity. This may be done as follows \begin{equation}
\Pr\left(y|\mathbf{x}\right)=\frac{Q\left(y|\mathbf{x}\right)}{\sum_{y^{\prime}=1}^{M}Q\left(y^{\prime}|\mathbf{x}\right)}\label{Eq:Probability}\end{equation}
where $Q\left(y|\mathbf{x}\right)\geq0$ (note that there is a slight
change of notation compared with \cite{Luttrell1997}, because $Q\left(y|\mathbf{x}\right)$
rather than $Q\left(\mathbf{x}|y\right)$ is written, but the results
are equivalent). The $Q\left(y|\mathbf{x}\right)$ are thus unnormalised
probabilities, and $\sum_{y^{\prime}=1}^{M}Q\left(y^{\prime}|\mathbf{x}\right)$
is the normalisation factor. The derivative of $\Pr\left(y|\mathbf{x}\right)$
is given by \begin{equation}
\frac{\delta\Pr\left(y|\mathbf{x}\right)}{\Pr\left(y|\mathbf{x}\right)}=\frac{1}{Q\left(y|\mathbf{x}\right)}\left(\delta Q\left(y|\mathbf{x}\right)-\Pr\left(y|\mathbf{x}\right)\sum_{y^{\prime}=1}^{M}\delta Q\left(y^{\prime}|\mathbf{x}\right)\right)\label{Eq:DerivativeProbability}\end{equation}

The second step in modelling $\Pr\left(y|\mathbf{x}\right)$ is to
introduce an explicit parameteric form for $Q\left(y|\mathbf{x}\right)$.
The following sigmoidal function will be used in this paper \begin{equation}
Q\left(y|\mathbf{x}\right)=\frac{1}{1+\exp\left(-\mathbf{w}\left(y\right)\cdot\mathbf{x}-b\left(y\right)\right)}\label{Eq:Sigmoid}\end{equation}
where $\mathbf{w}\left(y\right)$ is a weight vector and $b\left(y\right)$
is a bias. The derivative of $Q\left(y|\mathbf{x}\right)$ is given
by \begin{equation}
\delta Q\left(y|\mathbf{x}\right)=Q\left(y|\mathbf{x}\right)\left(1-Q\left(y|\mathbf{x}\right)\right)\left(\delta\mathbf{w}\left(y\right)\cdot\mathbf{x+w}\left(y\right)\cdot\delta\mathbf{x}+\delta b\left(y\right)\right)\label{Eq:DerivativeSigmoid}\end{equation}

\bigskip{}
There are also $\delta\mathbf{x}$ derivatives in equation \ref{Eq:DerivativeD1D2}
and equation \ref{Eq:DerivativeSigmoid}. The $\delta\mathbf{x}$
derivative arises only in multi-stage chains of FMCs, and because
of the way in which stages of the chain are linked together (see equation
\ref{Eq:ChainLink}) it is equal to the derivative of the vector of
probabilities output by the previous stage. Thus the $\delta\mathbf{x}$
derivative may be obtained by following its dependencies back through
the stages of the chain until the first layer is reached; this is
essentially the chain rule of differentiation. This ensures that for
each stage the partial derivatives include the additional contributions
that arise from forward propagation through later stages, as described
in section \ref{Sect:TrainingAlgorithm}.

There are also $\delta\mathbf{x}^{\prime}\left(y\right)$ derivatives
in equation \ref{Eq:DerivativeD1D2}, but these require no further
simplification.

\end{document}